\pdfoutput=1

\documentclass[11pt]{article}

\usepackage{ACL2023}

\usepackage{times}
\usepackage{latexsym}
\usepackage{multirow}
\usepackage{todonotes}
\usepackage{comment}
\usepackage[multiple]{footmisc}

\usepackage[T1]{fontenc}

\usepackage[utf8]{inputenc}

\usepackage{microtype}
\usepackage{xcolor}
\usepackage{inconsolata}

\usepackage{pifont}
\usepackage[normalem]{ulem}

\title{Do I have the Knowledge to Answer? \\ Investigating Answerability of Knowledge Base Questions}

\author{First Author \\
  Affiliation / Address line 1 \\
  Affiliation / Address line 2 \\
  Affiliation / Address line 3 \\
  \texttt{email@domain} \\\And
  Second Author \\
  Affiliation / Address line 1 \\
  Affiliation / Address line 2 \\
  Affiliation / Address line 3 \\
  \texttt{email@domain} \\}
\author{Mayur Patidar$^\dag$, Prayushi Faldu$^\ddag$, Avinash Singh$^\dag$, Lovekesh Vig$^\dag$,\\ {\bf Indrajit Bhattacharya$^\dag$, } {\bf Mausam$^\ddag$}\\
$^\dag$TCS Research,
$^\ddag$Indian Institute of Technology, Delhi \\
\{patidar.mayur, singh.avinash9, lovekesh.vig, b.indrajit\}
@tcs.com
\,\\ 
prayushifaldu123@gmail.com, mausam@cse.iitd.ac.in
}

\begin{document}
\maketitle
\newcommand{\olddata}{{ GrailQA}}
\newcommand{\data}{{ GrailQAbility}}

\begin{abstract}
When answering natural language questions over knowledge bases, 
missing facts, incomplete schema and limited scope naturally lead to many questions being unanswerable. 
While answerability has been explored in other QA settings, 
it has not been studied for QA over knowledge bases (KBQA). 
We create \emph{\data}, a new benchmark KBQA dataset with unanswerability, by first identifying various forms of KB incompleteness that make questions unanswerable, and then systematically adapting \olddata\ (a popular KBQA dataset with only answerable questions). 
Experimenting with three state-of-the-art KBQA models, 
we find that all three models suffer a drop in performance even after suitable adaptation for unanswerable questions. 
In addition, these often detect unanswerability for wrong reasons and find specific forms of unanswerability particularly difficult to handle.
This underscores the need for further research in making KBQA systems robust to unanswerability.

\end{abstract}

\section{Introduction}\label{sec:intro}

The problem of natural language question answering over knowledge bases (KBQA) has received a lot of interest in recent years ~\cite{saxena:acl2020kgqaembed,zhang:acl2022subgraph,mitra:naacl2022cmhopkgqa,wang:naacl2022mhopkgqa,das:icml2022subgraphcbr,cao:acl2022progxfer,ye:acl2022rngkbqa,chen:acl2021retrack,das:emnlp2021cbr}.
An important aspect of this task for real-world deployment is detecting answerability of questions. 
This problem arises for KBs due to various reasons, including schema-level and data-level incompleteness of KBs~\cite{Min2013DistantSF}, limited KB scope, questions with false premises, etc.
In such cases, a robust and trustworthy model should detect and report that a question is unanswerable, instead of outputting some incorrect answer.

Answerability is well studied for QA over unstructured contexts~\cite{rajpurkar:acl2018-squadidk,choi:emnlp2018-quac,reddy:tacl2019coqa,sulem:naacl2022-ynidk,raina:acl2022-mcqnone}.
However, there is no existing work on answerability for KBQA.
Benchmark KBQA datasets~\cite{gu:www2021grailqa,yih:acl2016webqsp,talmor:acl2018complexwebqsp,cao:acl2022kqapro} contain only answerable questions.


\begin{figure*}[t]
\centering
\includegraphics[width=0.99\textwidth]{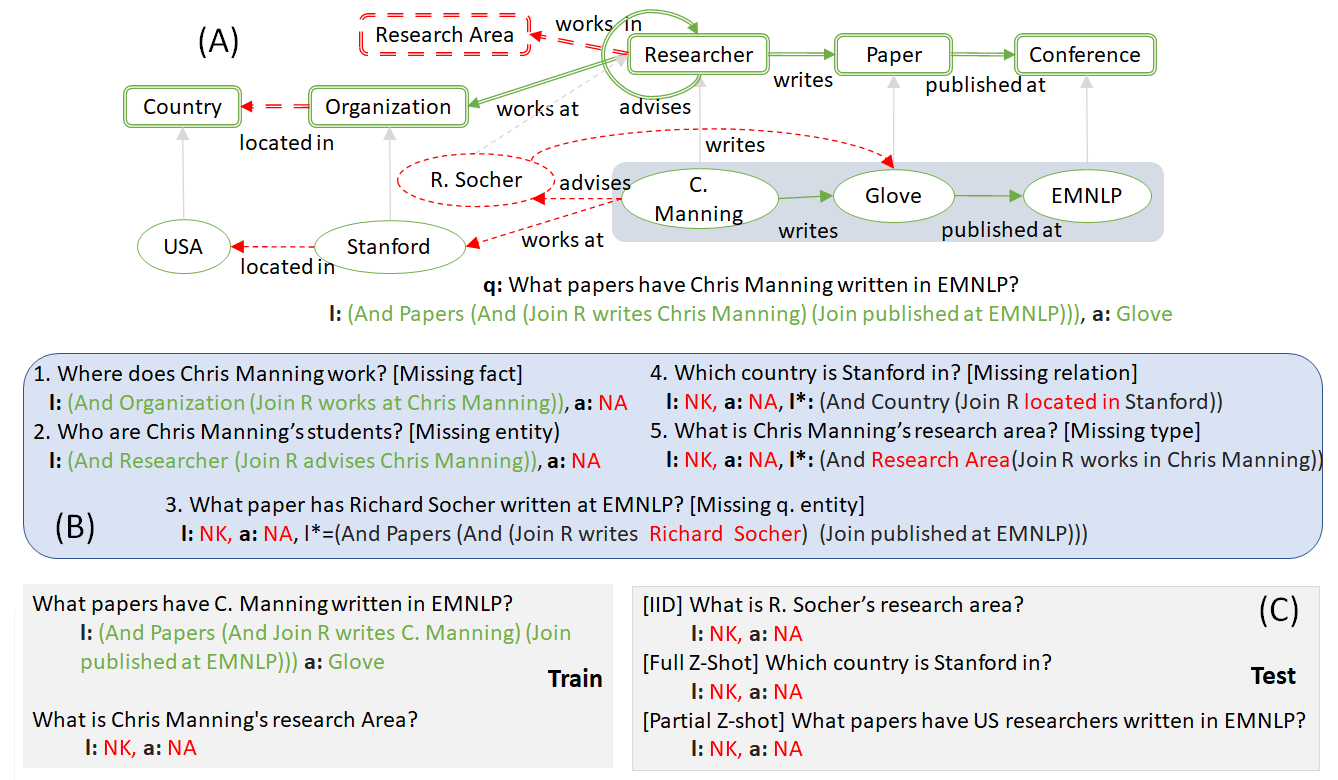}  
\caption{
(A) KB schema and facts. Elements in red are part of `ideal' KB but missing in the given KB for QA. An answerable question is shown for this KB with logical form $l$ in s-expression, answer $a$ and path (shaded blue).
(B) 5 types of unanswerable questions for provided KB, with actual logical forms $l$, answers $a$ and ideal logical forms $l^*$ with missing KB elements in red (3-5).
(C) Illustration of 3 different types of unanswerability scenario in test.
}  
\label{fig:example}
\end{figure*}

We first identify how different categories of KB incompleteness (schema and data incompleteness) affect answerability of questions.  
Then, using GrailQA~\cite{gu:www2021grailqa}, one of the largest KBQA benchmark dataset, we create a new benchmark for KBQA with unanswerable questions, which we call \emph{GrailQAbility}, by deleting various elements from the KB to simulate scope and fact coverage limitations.
This involves addressing a host of challenges, 
arising due to different ways in which KB element deletion affects answerability of questions, dependence between deletion of different types of KB elements, the shared nature of KB elements across questions, and more.
We also define and include different generalization scenarios for unanswerable questions in the test set, namely IID and zero-shot, mirroring those for answerable questions.



We then use GrailQAbility to evaluate the robustness of three recent state-of-the-art KBQA models, RnG-KBQA~\cite{ye:acl2022rngkbqa}, ReTraCk~\cite{chen:acl2021retrack} and TIARA~\cite{shu-etal-2022-tiara}, against unanswerable KB questions.
We find that all three models suffer an overall drop in performance with unanswerable questions, even after appropriate adaptation for unanswerability via retraining and thresholding.
More alarmingly, these often detect unanswerability for incorrect reasons, raising concerns about trustworthiness. 
Additionally, while the strength of these models is that they learn at the schema-level, we find that this also results in significantly poorer ability to detect data-level incompleteness. 
Using error analysis, we identify important failure points for these models. 
All of these highlight robustness issues for KBQA models in real applications, raising important questions for future research.

In summary, our contributions are as follows. (a) We motivate and introduce the task of detecting answerabilty for KBQA. (b) We create GrailQAbility, which is the first benchmark for KBQA with unanswerable questions. 
(c) Using experiments and analysis on GrailQAbility with 
three state-of-the-art KBQA models
, we identify aspects of unanswerability that these models struggle to identify.
We release code and data for further research.\footnote{\url{https://github.com/dair-iitd/GrailQAbility.git}\label{codefootnote}}



\section{KBQA with Answerability Detection}
\label{sec:task}

A {\em Knowledge Base} (KB) (also called Knowledge Graph) $G$ contains a {\em schema} $S$ (or ontology) with {\em entity types} (or types) $T$ and {\em relations} $R$ defined over pairs of types, which we together refer to as {\em schema elements} of the KB.
The types in $T$ are often organized as a hierarchy.
It also contains {\em entities} $E$ as instances of types, and {\em facts} (or triples) $F\subseteq E \times R \times E$, which we together refer to as {\em data elements} of the KB.
The top layer of Fig.~\ref{fig:example}(A) shows example schema elements, while the bottom layer shows entities and facts.
In {\em Knowledge Base Question Answering} (KBQA), we are given a {\em question} $q$ written in natural language which needs to be translated to a {\em logical form} (or query) $l$ that executes over $G$ to yield a set of {\em answers} $A$.
Different logical forms, SPARQL~\cite{webqsp}, s-expressions~\cite{gu:www2021grailqa}, programs~\cite{KQAPro}, etc., have been used in the KBQA literature.
We concentrate on {\em s-expressions}~\cite{gu:www2021grailqa}, 
which employ set-based semantics and functions with arguments and return values as sets.
These can be easily translated to KB query languages such as SPARQL, and provide a balance between readability and compactness~\cite{gu:www2021grailqa}.
We call a logical form {\em valid} for $G$ if it executes over $G$ without an error.
On successful execution, a logical form traces a {\em path} in the KB leading to each answer.
Fig.~\ref{fig:example}(A) shows an example query with a valid logical form (using s-expression) and the path traced by its execution.

We define a question $q$ to be {\bf answerable} for a KB $G$, if {\bf (a)} $q$ admits a valid logical form $l$ for $G$, AND {\bf (b)} $l$ returns a {\em non-empty} answer set $A$ when executed over $G$.
The example question in Fig.~\ref{fig:example}(A) is answerable for the shown KB.
The \emph{standard} KBQA task over a KB $G$ is to output the answer $A$, and optionally the logical form $l$, given a question $q$, {\em assuming $q$ to be answerable for $G$}.

Most recent KBQA models~\cite{ye:acl2022rngkbqa,chen:acl2021retrack} are trained with questions and gold logical forms.
Other models directly generate the answer~\cite{sun-etal-2019-pullnet,saxena:acl2022seq2seqembedkgqa}.
Different train-test settings
have been explored and are included in benchmark KBQA datasets~\cite{gu:www2021grailqa}.
For a question $q$, let $S_q$ denote the schema elements in the logical form for $q$.
Given a training set $Q_{tr}$, a test question $q$ is labelled {\em iid} if it follows the distribution for questions in $Q_{tr}$, and contains only schema elements seen in train $S_q\subseteq S_{Q_{tr}}$ (we have overloaded notation to define $S_{Q_{tr}}$).
Alternatively, a test question $q$ is labelled {\em zero shot} if it involves at least one unseen schema element ($S_q\not \subseteq S_{Q_{tr}}$).
Finally, test question $q$ involves {\em compositional generalization} if $S_q\subseteq S_{Q_{tr}}$ but the specific logical form for $q$ does not match that for any $q'\in Q_{tr}$.

By negating the above answerability definition, we define a question $q$ to be {\bf unanswerable} for a KB $G$ if {\bf (a)} $q$ does not admit a valid logical form $l$ for $G$, or {\bf (b)} the valid $l$ when executed over the $G$ returns an empty answer. 
Clearly, meaningless and out-of-scope questions for a KB are unanswerable.
Even for a meaningful question, unanswerability arises due to incompleteness (in data or schema) in $G$.
Such questions admit an `ideal KB' $G^*$ for which $q$ has a valid ideal logical form $l^*$ which executes on $G^*$ to generate a non-empty ideal answer $a^*$. 
The available KB $G$ lacks one or more schema or data elements making $q$ unanswerable.
Fig.~\ref{fig:example}(A) illustrates an available KB, with missing elements with respect to the ideal KB shown in red.
In Fig.~\ref{fig:example}(B), questions 1-2 yield valid queries for the available KB but missing facts lead to empty answers, while questions 3-5 lack schema elements for valid queries.

The task of {\bf KBQA with answerability detection}, given a question $q$ and an available KB $G$, is to {\bf (a)} appropriately label the answer $A$ as $\mbox{NA}$ (No Answer) or the logical form $l$ as $\mbox{NK}$ (No Knowledge, i.e., query not possible) when $q$ is unanswerable for $G$, or {\bf (b)} generate the correct non-empty answer $A$ and valid logical form $l$ when $q$ is answerable for $G$.
The training set may now additionally include unanswerable questions labeled appropriately with $A=\mbox{NA}$ or $l=\mbox{NK}$.
Note that training instances do not contain `ideal' logical forms for the unanswerable questions that have $l=\mbox{NK}$.

Mirroring answerable questions, we define different train-test scenarios for unanswerable questions as well. 
An {\em iid unanswerable} question in test follows the same distribution as unanswerable questions in train, and all missing KB elements (schema elements in its ideal logical form 
and missing data elements in its ideal paths)
are encountered in train unanswerable questions associated with the same category of incompleteness.
For example, the missing schema element {\em Research Area} for the first test question in Fig.~\ref{fig:example}(C) is covered by the second train question.
In contrast, a {\em zero-shot unanswerable test question} involves at least one missing KB element (schema element in its ideal logical form or data element in its paths) that is not part of any unanswerable question in train associated with same category of incompleteness.
E.g., the missing schema elements ({\em located in} and {\em works at}) for the second and third test questions in Fig.\ref{fig:example}(C) are not covered by any unanswerable question in train.
We further define two sub-classes, partial and complete zero-shot, for zero-shot unanswerable questions, but for clarity, discuss these in Sec. \ref{sec:results}.

\section{GrailQAbility: Extending GrailQA with Answerability Detection}
\label{sec:dataset}

\begin{table*}[h]
        \begin{center}
            \normalsize
            \begin{tabular}[b]{|c|cccccc|cc|cc|}
\hline
\multirow{2}{*}{Dataset} &\multirow{2}{*}{\#Q} &\multirow{2}{*}{\#LF} &\multirow{2}{*}{\#D} &\multirow{2}{*}{\#R} &\multirow{2}{*}{\#T} &\multirow{2}{*}{\#E} &\multicolumn{2}{c|}{Q. Type} &\multicolumn{2}{c|}{Test Scenarios} 
  \\ 
    & & & & & &  &A & U&A  &U
  \\ 
\hline

 GrailQA &64,331 & 4969 & 86 & 3720 & 1534 & 32,585  &\ding{51} & \ding{55} & I, C, Z & \ding{55}   \\ 
  GrailQAbility &50,507 & 4165  & 81 & 2289 & 1081 & 22,193   &\ding{51} & \ding{51} & I, C, Z & I, Z  \\

  \hline 
\end{tabular}
\end{center}
\caption{
Statistics for GrailQA and GrailQAbility. \#Q is no. of questions, \#LF no. of unique canonical logical forms, \#D no. of domains, \#R, \#T, \#E no. of relations, types and entities, A and U denote answerable and unanswerable questions. I, C, and Z denote IID, compositional and zero-shot.}
\label{tab:dataset_comparison}
\end{table*}

In this section, we describe the creation of a new benchmark dataset for KBQA with unanswerable questions.
In a nutshell, we start with a standard KBQA dataset containing only answerable questions for a given KB.
We introduce unanswerability in steps, by deleting schema elements (entity types and relations) and data elements (entities and facts) from the given KB.
We mark questions that become unanswerable as a result of each deletion with appropriate unanswerability labels.
We control the percentage of questions that become unanswerable as a result of each type of deletion.

Many complications arise in this. 
(a) Deletion of different KB elements affect answerability differently. 
Some affect logical forms and answers, while others affect answers only.
(b) The same KB element potentially appears in paths or logical forms of multiple questions. 
(c) KB elements cannot be deleted independently -- entity types are associated with relations and entities, while relations and entities are associated with facts. 
(d) Questions with multiple answers remain answerable until the fact paths to {\em all} of these answers have been broken by deletions. 
(e) Choosing KB elements to delete uniformly at random does not resemble incompleteness in the real world.   

We address these issues as follows.
(a-b) We iterate over the 4 categories of KB elements to be deleted, efficiently identify affected questions for a deleted schema element using an index, tag these with the deleted type, and appropriately relabel their logical forms or answers. 
We stop when specific percentages of questions are unanswerable for each category.
(c) We delete different types of KB elements in an appropriate sequence -- entity types, followed by relations, entities and finally facts.
(d) We track remaining fact paths for questions and mark a question as unanswerable only when all paths are broken by KB deletions.
(e) When sampling KB elements to delete, since ``better known'' KB elements are less likely to be missing, we incorporate the inverse popularity of an element in the original KB in the sampling distribution.
Additionally, we only consider those elements present in still valid logical forms and paths for the questions in the dataset. 
Next, we describe the specifics for individual KB element categories. 


\vspace{0.5ex}
\noindent
\textbf{Fact Deletion:}
Dropping a KB fact can break the path of one or more answers for a question but cannot affect the logical form.
Answers whose paths are broken are removed from the answer list of the question.
If the answer list becomes empty as a result of a fact drop, we set its answer to $\mbox{NA}$ but leave its logical form unchanged.
In Fig.\ref{fig:example}(B), deleting ({\em C. Manning, works at, Stanford}) makes Q1 unanswerable.

\vspace{0.5ex}
\noindent
\textbf{Entity Deletion:}
To delete an entity from the KB, we first delete all its associated facts, and then drop the entity itself. 
Deleting facts affects answerability of questions as above, as for Q2 in Fig.\ref{fig:example}(B).
Deleting an entity additionally affects answerability of questions whose logical form contains that entity as one of the mentioned entities.
This happens for Q3 in Fig.\ref{fig:example}(B) when entity {\em R. Socher} is deleted.
For such questions, the logical form also becomes invalid, and we set it as $\mbox{NK}$.

\vspace{0.5ex}
\noindent
\textbf{Relation Deletion:}
To delete a relation, we first drop all facts associated with it, and then drop the relation itself from the schema. 
Deleting facts makes some questions unanswerable as above, and we set their answers to be $\mbox{NA}$. 
Deleting the relation additionally affects the logical form of some questions, and we set their logical forms to be $\mbox{NK}$.
This happens for Q4 in Fig.~\ref{fig:example}(B) on deleting the {\em located in} relation. 

\vspace{0.5ex}
\noindent
\textbf{Entity Type Deletion:}
Entities are often tagged with multiple types in a hierarchy (e.g {\em C. Manning} may be {\em Researcher} and {\em Person}).
After deleting an entity type from the KB schema, we also delete all entities $e$ that are associated {\em only} with that type.
We further delete all relations associated with the type.
For Q5 in Fig.\ref{fig:example}(b), the logical form becomes invalid on deleting the {\em Research Area} entity type.
For an affected question, we set its answer as $\mbox{NA}$ and its logical form as $\mbox{NK}$.

\begin{table}[h]
\begin{center}
\small
            \begin{tabular}[b]{|c|c|c|c|}
\hline
\multirow{2}{*}{Split} &\multirow{2}{*}{A}&\multicolumn{2}{c|}{U} \\
\cline{3-4}
 &  & NK & NA  \\ 
\hline
    
Train &23,933  &7110 &4240  \\ 
Dev &3399  &1064  &  595 \\ 
Test & 6808 &2162 &1196  \\ \hline
\end{tabular}
\end{center}
\caption{GrailQAbility: Train, Dev and Test Splits}
\label{tab:train_dev_test_split}
\end{table}

\paragraph{GrailQAbility Dataset: }
We make use of GrailQA~\cite{gu:www2021grailqa}, which is one of the largest and most diverse KBQA benchmark based on Freebase but contains only answerable questions, and create a new benchmark for KBQA with answerability detection.
We call this GrailQAbility (GrailQA with Answerability).
We make this dataset public.
Aligning with earlier QA datasets with unanswerability~\cite{rajpurkar:acl2018-squadidk,sulem:naacl2022-ynidk,choi:emnlp2018-quac,raina:acl2022-mcqnone}, we keep the total percentage of unanswerable questions as 33\%, splitting this nearly equally (8.25\%) between deleted entity types, relations, entities and facts. 

{\em Train-Test Split:}
Since the test questions for GrailQA are unavailable, we use the train and dev questions.
We keep aside the compositional and zero shot questions from dev as the compositional and zero shot {\em answerable} questions in our new dev and test set.
We then combine the train and iid dev questions, introduce unanswerability into these by running the 4 categories of deletion algorithms in sequence, and split these to form the new train and iid test+dev (both answerable and unanswerable) and zero shot unanswerable test+dev questions.
The unanswerable questions in test and dev contain 47\% iid, and 53\% zero-shot.
Statistics for GrailQAbility and GrailQA are compared in Tab.~\ref{tab:dataset_comparison}. 
Sizes of the different splits are shown in Tab.~\ref{tab:train_dev_test_split}.
Details on dataset creation are in appendix (\ref{subsec:details_of_datatset_creation}).

\section{Experimental Setup}
\label{sec:expt}

\begin{table*}[ht]
\begin{center}
\small
            \begin{tabular}[b]{|c|l|ccc|ccc|ccc|}
\hline

  \multicolumn{1}{|c}{Train}&\multicolumn{1}{|c}{Model}&\multicolumn{3}{|c}{Overall}
  & \multicolumn{3}{|c}{Answerable}  & \multicolumn{3}{|c|}{Unanswerable}  \\

 & & F1(L) & F1(R) & EM & F1(L) & F1(R) & EM & F1(L) & F1(R) & EM \\ \hline
  \multirow{6}{*}{A}&RnG-KBQA & 67.8 &65.6  &51.6  &  78.1  &  78.1  & 74.2  &46.9  &40.1  & 5.7 \\ 
 &RnG-KBQA+T & 67.6 &65.8  &57.0  & 71.4 &  71.3& 68.5 & 59.9 &54.5  &33.6  \\ 
  &ReTraCk &69.2 &67.0 &50.7  &67.0  &66.9  &62.4  &73.8  &67.2  &27.1  \\

   &ReTraCk+T &69.9  &67.9  &52.0  &65.3  &65.3  &61.2 &79.3  &73.2  &33.4  \\
   &TIARA &77.1 &75.0 &56.0  &{\bf82.9}  &{\bf82.8}  &{\bf79.2}  &65.4  &59.0  &9.0  \\
   &TIARA+T & 76.5&74.8 &63.4  &76.9  & 76.8 & 74.1 & 75.9 &70.8  &41.8  \\
  \hline 
  \multirow{6}{*}{A+U}&RnG-KBQA &  80.5 &  79.4 &  68.2  &75.9  &75.9  &72.6  &89.7  &86.4  &59.4  \\ 
 &RnG-KBQA+T & 77.8 &77.1  &67.8  &70.9  &70.8  & 68.1 &  92.0  &  89.8  &  67.2  \\ 
  &ReTraCk &69.7  &68.4  &56.5  &61.4  &61.3  &57.3  &86.5  &82.8  &54.7  \\
  &ReTraCk+T &70.3  &69.1 &56.6  &61.2  &61.1  &57.1  &88.7  &85.1  &55.5  \\
  &TIARA & {\bf83.9}&{\bf82.9} &69.7  &81.0  &81.0  & 78.3 & 89.9 &86.8  &52.3  \\
   &TIARA+T & 81.7& 81.1&{\bf72.6}  &76.0  &76.0  &74.0  &{\bf93.3}  &{\bf91.3}  &{\bf69.8}  \\
  \hline 
\end{tabular}
\end{center}
\caption{Performance of different models on GrailQAbility over all, answerable and unanswerable questions. EM is exact match on logical forms and F1(L) and F1(R) are lenient and regular evaluations of answers. A and A+U indicate training with only answerable questions and with both answerable and unanswerable questions. Models with suffix +T have additional thresholds for entity disambiguation and logical form fine-tuned on dev set.}
\label{tab:mainresult}
\end{table*}

\begin{table*}[ht]
        \begin{center}
            \small
            \begin{tabular}[b]{|c|l|cccccc|cccc|}
\hline
  \multicolumn{1}{|c}{Train}&\multicolumn{1}{|c}{Model}&\multicolumn{6}{|c}{Schema Element Missing}  
  & \multicolumn{4}{|c|}{Data Element Missing}  \\     
  & &\multicolumn{2}{c}{Type}  
  & \multicolumn{2}{c}{Relation}  & \multicolumn{2}{c|}{Mention Entity} &
  \multicolumn{2}{c}{Other Entity} &
  \multicolumn{2}{c|}{Fact}  \\     
  & & F1(R) & EM &  F1(R) & EM  & F1(R) & EM &F1(R) & EM & F1(R) & EM  \\ \hline
  \multirow{6}{*}{A}&{\small RnG-KBQA} & 40.1  &0.0    &44.2  & 0.0   &27.4  &0.0 &45.1  & 13.5   &46.0  &16.8 \\ 
 &{\small RnG-KBQA+T}  &55.5  &49.5    &57.1  &46.6    &44.7  & 40.3  &56.0  & 11.5& 58.6  &13.9 \\ 
  &{\small ReTraCk}   &71.1  &34.8    &59.3  &18.9    &80.7  &63.7 &72.6  &11.2   &64.4  &11.9 \\

   &{\small ReTraCk+T}   &75.7  &47.9 &64.9  &28.8    &83.5  &70.3 &81.0  &10.9   &72.3  &12.0 \\
   &{\small TIARA}   & 57.7 &0.0    &56.9  &0.0    &51.9  &0.0 & 65.8 &22.4   &  65.8& 26.3\\
   &{\small TIARA+T}   & 68.0 &56.5    &69.5  &48.7    &74.4  &62.6 &70.9  &18.5   &74.0  &20.9 \\
  \hline 
  \multirow{6}{*}{A+U}&{\small RnG-KBQA}  &91.6  &75.8& 86.4 & 66.6   &87.6  &72.0&84.0  & 37.5   &82.4  & 39.1  \\ 
 &{\small RnG-KBQA+T}   & {\bf93.4}  &  {\bf86.8}    & 89.7  &  {\bf85.5}    &  92.1  &  {\bf89.6} & 87.1  &30.8  &  86.0  &32.5 \\ 
  &{\small ReTraCk}   &89.6  &82.2    &86.4  &74.4    &90.3  &85.9   &79.0  &9.8    &71.7  &10.8 \\
  &{\small ReTraCk+T}   &90.6  &83.1    &87.8  &76.0    &91.2 &86.8 &83.2  &9.8 &76.4  &10.8 \\
  &{\small TIARA}   & 83.7 &50.6    &83.6  &40.5    &88.7  &52.5 & {\bf91.6} &{\bf62.5}   &90.9  &{\bf63.7} \\
   &{\small TIARA+T}   &88.9  &80.3    &{\bf90.9}  &77.1    & {\bf94.7} &84.6 &91.6  &53.2   &{\bf92.6}  &53.4 \\
  \hline 
\end{tabular}
\end{center}
\caption{Performance for different KBQA models for subsets of questions affected by different types of KB incompleteness. 
Note that missing mention entities result in invalid logical form and other missing entities lead to valid logical form with no answer. 
Names have the same meanings as in Tab.~\ref{tab:mainresult}.}
\label{tab:drop_type}
\end{table*}

\begin{table}[h]
\begin{center}
\small            
\begin{tabular}[b]{|c|l|cc|cc|}
\hline
  \multicolumn{1}{|c}{Train}&\multicolumn{1}{|c}{Model}&\multicolumn{2}{|c}{IID}
  
  & \multicolumn{2}{|c|}{Zero-Shot }    \\

 & & F1(R) & EM & F1(R) & EM  \\ \hline

  \multirow{5}{*}{A+U}&RnG-KBQA &91.9 &73.3 &81.7 &47.1   \\ 
 &RnG-KBQA+T &  94.3 &  75.9 &  85.9 &  59.5   \\ 
  &ReTraCk &88.7 &66.5 &77.7 &44.4   \\
  &ReTraCk+T &90.1 &66.6 &80.7 &45.7 \\
  &TIARA & 90.9&63.4 &83.1 &42.5   \\
  &TIARA+T &{\bf94.5} & {\bf78.7}& {\bf88.5}&{\bf62.0} \\
  \hline 
\end{tabular}
\end{center}
\caption{Performance of different models for unanswerable IID and zero-shot test scenarios in GrailQAbility. Names have the same meanings as in Tab.~\ref{tab:mainresult}.}
\label{tab:test_scenarios}
\end{table}

\begin{table}[ht]
\begin{center}
\small            
\begin{tabular}[b]{|c|l|cc|cc|}
\hline
  \multicolumn{1}{|c}{Train}&\multicolumn{1}{|c}{Model}
  
  & \multicolumn{2}{|c}{Full Z-Shot }  & \multicolumn{2}{|c|}{Partial Z-Shot}  \\

 &  & F1(R) & EM & F1(R) & EM \\ \hline
  \multirow{5}{*}{A+U}&RnG-KBQA     &87.2  &75.9 &78.0  &40.0 \\ 
 &RnG-KBQA+T    &  89.7  & {\bf 86.7} & {\bf 83.1}  & {\bf 71.0}   \\ 
  &ReTraCk     &86.2  &65.0 &73.6  &54.5  \\
  &ReTraCk+T      &88.2  &67.0  &75.6  &56.7 \\
  &TIARA     & 85.7 & 41.9&73.7   &20.0   \\
  &TIARA+T      &{\bf90.6}   &72.4  &82.0  &64.0 \\
  \hline 
\end{tabular}
\end{center}
\caption{Performance of different models for partial zero-shot and full-zero test scenarios in GrailQAbility. Names have the same meanings as in Tab.~\ref{tab:mainresult}.}
\label{tab:zeroshot_drilldown}
\end{table}

\paragraph{KBQA Models:}
Among state-of-the-art KBQA models, we pick RnG-KBQA~\cite{ye:acl2022rngkbqa}, ReTraCk~\cite{chen:acl2021retrack} and TIARA~\cite{shu-etal-2022-tiara}.
These report state-of-the-art results on GrailQA as well as on WebQSP~\cite{berant-etal-2013-semantic,yih:acl2016webqsp,talmor:acl2018complexwebqsp} - the two main benchmarks.
On the GrailQA leader board,\footnote{\url{https://dki-lab.github.io/GrailQA/}} 
these are the top three published models with available code (at the time of submission).
Since these generate logical forms, we expect these to be more robust to data level incompleteness than purely retrieval-based approaches~\cite{saxena:acl2020kgqaembed,das:emnlp2021cbr,zhang:acl2022subgraph,mitra:naacl2022cmhopkgqa,wang:naacl2022mhopkgqa}. 

{\bf RnG-KBQA}~\cite{ye:acl2022rngkbqa} 
first uses a BERT-based ~\cite{Devlin2019BERTPO} ranker to select a set of candidate logical forms for a question by searching the KB, and then a T5-based ~\cite{2020t5} model generates the logical form using the question and candidates.  
{\bf ReTraCk}~\cite{chen:acl2021retrack} also uses a rank and generate approach, but uses a dense retriever to retrieve schema elements for a question, and grammar-guided decoding using an LSTM ~\cite{lstm} to generate the logical form using the question and retrieved schema items.
{\bf TIARA}~\cite{shu-etal-2022-tiara} combines the retrieval mechanisms of the first two models to include both candidate logical forms as well as candidate schema elements from the KB.
It then uses constrained decoding like ReTraCk but using T5~\cite{raffel:jmlr2020-t5}.
All three models use entity disambiguation to find KB entities mentioned in a question and also check execution for generated logical forms.

\paragraph{Adapting for Answerability:} 
We use existing code bases\footnote{\url{https://github.com/salesforce/rng-kbqa}}\footnote{\url{https://github.com/microsoft/KC/tree/main/papers/ReTraCk}} \footnote{\url{ https://github.com/microsoft/KC/
tree/main/papers/TIARA}} 
of these models, and adapt these in two ways --- thresholding and training with unanswerability.
ReTraCk 
and TIARA
return empty logical form when execution fails, which we interpret as $l=\mbox{NK}$ prediction.
For all models, we additionally introduce thresholds for entity disambiguation and logical form generation, and take the prediction to be $\mbox{NK}$ when the scores for entity linking and logical form are less than their corresponding thresholds. 
These thresholds are tuned using the validation set.
We train the models as in their original setting with only the answerable subset of training questions, leaving out the unanswerable questions ({\bf A training}).
We also train by including both the answerable and unanswerable questions in the training data ({\bf A+U training}).
More details are in appendix (\ref{sec:traindetails}).

\paragraph{Evaluation Measures: }
To evaluate a model's performance for detecting unanswerability, we primarily focus on the correctness of the logical form. 
We compare the predicted logical form with the gold-standard one using exact match (EM)~\cite{ye:acl2022rngkbqa}.
As it is ultimately a QA task (and other systems may produce answers without generating logical forms), we also perform direct answer evaluation.
Since in general a question may have multiple answers, we evaluate predicted answers using precision, recall and F1. 
In regular answer evaluation (R), we compare the predicted answer (which could be NA) with the gold answer in the modified KB, as usual.
Specifically for unanswerability, we also consider lenient answer evaluation (L), where we account for the gold answer in the original (ideal) KB as well, and also give credit to models which are able to recover this answer, perhaps via inference.
As an example, for the second test question in Fig.~\ref{fig:example}(C), R-evaluation only rewards $\mbox{NA}$ as answer, whereas L-evaluation rewards both $\mbox{NA}$ and {\em USA} as perfect answers.
Details of evaluation measures are in appendix (\ref{sec:lenientevaluation}).


\section{Results and Discussion} \label{sec:results}

We structure our discussion of experimental results around four research questions.

\paragraph{RQ1.} {\bf How do state-of-the-art KBQA models perform for answerability detection?}

Tab.~\ref{tab:mainresult} shows high-level performance for the three models on answerable and unanswerable questions.
We observe the following.

{\bf (A)} When training with only answerable questions (A training), all models perform poorly for unanswerable questions in terms of EM, ReTraCk being better than the other two.  

{\bf (B)} Performance improves for unanswerable questions with thresholding and A+U training but remains below the skyline for answerable questions with A training. 
The gap is $\sim$7 pct points for RnG-KBQA and ReTraCk and $\sim$9 for TIARA.

{\bf (C)} Not surprisingly, improvement for unanswerable questions comes at the expense of answerable question performance. 
The best overall performance (72.6 EM for TIARA) is $\sim$6.5 percentage points lower than the best answerable performance (79.2 EM for TIARA). 
Further, we observed that answerable performance is affected by thresholding (across iid, compositional and zero-shot settings) for all models.
This is also the case for the A+U training in the zero-shot setting. 
More details can be found in Tab.\ref{tab:test_scenarios_answerable} in appendix.
The reason is that for both forms of adaptation, the models incorrectly predict $l=\mbox{NK}$ for answerable questions. 

{\bf (D)} Unlike for answerable questions, there is a very large gap between EM and F1(R) for unanswerable questions. 
This is because correct NA (no answer) predictions are often associated with spurious logical form predictions, for all three models but for different reasons. 
We discuss this further under RQ4.

{\bf (E)} Performance is better (by about 2-4 percentage points) with lenient answer evaluation than with the regular counterpart. 
We found that this is often because the models generate logical forms with schema elements similar to the deleted ones, and return as a result subsets or supersets of the old answer instead of NA.
As one example, the question {\em Which football leagues share the same football league system as Highland Football League?} has 7 answers, but becomes unanswerable when the relation {\em soccer.football\_league\_system.leagues} is missing.
The model answers a different question - {\em Which football leagues play the same sport as Highland Football League} - by substituting the missing relation with {\em sports.sports\_league.sport}, and retrieves 152 answers, one of which, {\em Scottish Premier League}, is also in the original answer.

\paragraph{RQ2.} {\bf Are different forms of KB incompleteness equally challenging?} 

In Tab.~\ref{tab:drop_type}, we break down performance for unanswerable questions according to different forms of KB incompleteness.
Note that we have decomposed entity deletions further into deletion of mentioned entities (which affect the logical form) and other entities in the path (which affect only the answer paths).
The following are the main takeaways.

{\bf (A)} Performance (EM) is significantly poorer for all forms of missing data elements than missing schema elements, even after thresholding and retraining.
TIARA is an exception and  performs better for missing data elements with A+U training. 

{\bf (B)} A+U training significantly boosts performance for missing schema elements but not for missing data elements.
This is because RnG-KBQA and ReTraCk learn to generate logical forms involving schema elements. 
As a result, schema-level patterns are easier to learn 
for unanswerable questions with missing schema elements than those with missing data elements.
Secondly, these two rely on retrieved data paths to generate logical forms.
When relevant data elements are missing, the models fail to retrieve any familiar  input pattern and predict $l=\mbox{NK}$.
The interesting exception is TIARA.
By virtue of generating logical forms conditioned on both retrieved paths and schema elements and removing data path constraints during decoding, it learns to generate correct logical forms for missing data elements. 
But this also leads to the generation of syntactically valid but incorrect logical forms for missing schema elements.
However, these typically have low score and performance for missing schema elements improves with thresholding.

{\bf (C)} Gap between EM and F1(R) is small for missing schema elements ($l=\mbox{NK}$) and extremely large for missing data elements ($l\neq\mbox{NK}$), with the exception of A+U trained TIARA. 
Also, thresholding hurts performance for missing data elements.
This is because questions with missing data elements have valid logical forms,
and thresholding and A+U training produce $l=\mbox{NK}$ predictions which are themselves incorrect but imply $A=\mbox{NA}$ which is correct.
Thus we get correct $A=\mbox{NA}$ predictions for the wrong reason.

\paragraph{RQ3.} {\bf How difficult is zero shot generalization compared to iid for unanswerable questions?}

Recall that a zero-shot unanswerable test instance involves one or more missing KB elements that are not encountered in any unanswerable train instance with the same category of incompleteness.
Note that the definitions of iid and zero-shot make use of unanswerable training instances, so that only A+U training makes sense for this comparison.

{\bf (A)} The decomposition of unanswerable performance in terms of iid and zero-shot subsets is shown in  Tab.~\ref{tab:test_scenarios}.
As expected, iid performance is better than zero-shot for all models. 
The best performance is for TIARA+T (EM 78.7 for iid, 62 for zero-shot) which is marginally better than RnG-KBQA+T.

{\bf (B)} However, more interesting insights arise for unanswerability from a deeper drill-down of zero-shot instances.
We define a zero-shot instance to be {\em full zero-shot} when it does not involve any schema element seen in logical forms of answerable questions in train.
The second test question in Fig.~\ref{fig:example}(C), involving the missing relation {\em located in} is an example.
In contrast, a {\em partial zero-shot} unanswerable question is part ``seen answerable'' in addition to being part ``unseen unanswerable''. 
Specifically, its ideal logical form also contains at least one schema element seen for answerable questions in train.
The third test question in Fig.~\ref{fig:example}(C) is an example.
The {\em located in} and {\em works at} relations are ``new unseen'', while {\em writes} and {\em published at} are ``seen'' in the first train question, which is answerable.
Tab.\ref{tab:zeroshot_drilldown} shows full zero-shot and partial zero-shot performance for unanswerable questions. 
We see that all models find full-zero-shot to be significantly easier than partial zero-shot.
For RnG-KBQA+T, which is the best model, there is a 15.7 percentage point difference in EM. 
The reason is that partial zero-shot unanswerable questions have some KB elements seen during training (in answerable contexts), and some zero-shot KB elements (that make the question unanswerable) unseen during training. 
This confuses the models, which often labels these as answerable.
The full zero-shot instances do not have any similarity with training answerable questions and are less confusing.

We have not considered compositional generalization for unanswerable questions.
We may define a compositional unanswerable question as one that contains more than one missing KB element in its ideal logical form or in its ideal paths, all of which have appeared in unanswerable training instances, but not all in the same instance.
We hypothesize that detecting unanswerability in this scenario should only be hard as for IID unanswerability.
We plan to validate this experimentally in the future.
Additionally, since missing data elements constitute an important aspect of unanswerability for KB questions, we have included missing data elements in our definitions of iid and zero-shot unanswerability.
However, distributions at the level of KB data elements cannot realistically be learnt.
Therefore alternative definitions for these based only on schema elements may be more practical.

\paragraph{RQ4.} {\bf How do RnG-KBQA, ReTraCk and TIARA compare for unanswerable questions?}

On GrailQA (answerable questions with A-training), RnG-KBQA outperforms ReTraCk~\cite{ye:acl2022rngkbqa} and TIARA outperforms both~\cite{shu-etal-2022-tiara}, and we see the same pattern in GrailQAbility.
In the context of unanswerable questions, we make the following observations.

{\bf (A)} RnG-KBQA outperforms ReTraCk with thresholding and retraining by a similar margin as for answerable questions (12 pct points).
However, TIARA outperforms RnG-KBQA by a much smaller margin for unanswerable questions (2.6 pct points) compared to answerable ones (5 pct points).

{\bf (B)} With just A training, ReTraCk performs better than the other two models for unanswerable questions. 
This is due to the difference in fallback strategies when execution fails for generated logical forms.
ReTraCk's fallback acknowledges unanswerability ---
it returns empty logical form.
On the other hand, RnG-KBQA's fallback assumes answerability.
It returns logical forms corresponding to top-ranked paths or the nearest neighbor in the training set.
In settings with unanswerability, ReTraCk naturally performs better.
TIARA also has the ability to return empty logical forms, but this happens rarely --- when execution fails for generated logical forms and additionally the ranker output is  empty (i.e. no enumerations)).

{\bf (C)} We find that all models generate spurious logical forms, but for different reasons.
RNG-KBQA hallucinates relations that do not exist in the KB.
For example, when the relation {\em cricket\_tournament\_event.tournament} is deleted, RnG-KBQA substitutes that with the imaginary relationship {\em cricket\_tournament\_event.championship}.
ReTraCk and TIARA avoid this by virtue of constrained decoding, but incorrectly replace missing relations with other semantically or lexically relevant relations for the same entity.
For example, for the question {\em Which ac power plug standard can handle more than 50 Hz?}, when the {\em mains\_power.ac\_frequency} relation is missing, ReTraCk incorrectly replaces that with {\em power\_plug\_standard.rated\_voltage}.

{\bf (D)} With A+U training and thresholding, ReTraCk performs almost at par with RnG-KBQA for missing schema elements. 
But it performs significantly worse for missing data elements, for which its performance is hurt by these adaptations.  
This is because ReTraCk's constrained decoding forces it to always generate $l=\mbox{NK}$ in the absence of valid answer paths, which cannot be alleviated by additional training. 
Using decoding with syntactic constraints, TIARA establishes the best balance between missing schema and data elements and outperforms the other two models by a huge margin for missing data elements.
However for missing schema elements RnG-KBQA is the best individual model outperforming TIARA by 5-8 pct points.

\section{Related Work}\label{sec:rw}

\paragraph{KBQA models:} There has been extensive research on KBQA in recent years. 
Retrieval based approaches~\cite{saxena:acl2020kgqaembed,zhang:acl2022subgraph,mitra:naacl2022cmhopkgqa,wang:naacl2022mhopkgqa,das:icml2022subgraphcbr} learn to identify paths in the KB starting from entities mentioned in the question, and then score and analyze these paths to directly retrieve the answer.
Query generation approaches~\cite{cao:acl2022progxfer,ye:acl2022rngkbqa,chen:acl2021retrack,das:emnlp2021cbr} learn to generate a logical form or a query (e.g in SPARQL) based on the question, which is then executed over the KB to obtain the answer.
Some of these retrieve KB elements first and then use these in addition to the query to generate the logical form~\cite{ye:acl2022rngkbqa,chen:acl2021retrack}. 
~\citet{cao:acl2022progxfer} first generate a KB independent program sketch and then fill in specific arguments by analyzing the KB.
All these models have so far only been evaluated for answerable questions.
There is work on improving accuracy of QA over incomplete KBs~\cite{thai:arxiv2022cbrikb,saxena:acl2020kgqaembed}, but these do not address answerability.

\vspace{1ex}
\noindent
\textbf{Answerability in QA:} 
Answerability has been explored for extractive QA~\cite{rajpurkar:acl2018-squadidk}, conversational QA~\cite{choi:emnlp2018-quac,reddy:tacl2019coqa}, boolean (Y/N) QA~\cite{sulem:naacl2022-ynidk} and  MCQ~\cite{raina:acl2022-mcqnone}.
While our work is motivated by these, the nature of unanswerable questions is very different for KBs compared to unstructured contexts.
Also, KBQA models work differently than other QA models. 
These retrieve paths and KB elements
to prepare the context for a question. 
Relevant context is then pieced together to generate a logical
query rather than the answer directly. 
We find that this makes them more prone to mistakes in the
face of unanswerability.

In very recent work on unanswerability for text-to SQL~\cite{wang:arxiv2022},
the underlying cause of answerability --- missing columns in individual tables --- is much simpler than for KBQA. 

\vspace{1ex}
\noindent
\textbf{QA Datasets and Answerability:} Many benchmark datasets exist for KBQA~\cite{gu:www2021grailqa,yih:acl2016webqsp,talmor:acl2018complexwebqsp,cao:acl2022kqapro}, but only contain answerable questions. 
QALD~\cite{perevalov2:icsc022} is a multilingual dataset containing ``out-of-scope'' questions that may be considered unanswerable according to our definition. 
However, the number of such questions is very small (few tens in different versions of the dataset), which hinders any meaningful bench-marking. 
It also does not have any finer categorization of such questions.

Unanswerable questions have been incorporated  into other QA datasets~\cite{rajpurkar:acl2018-squadidk,sulem:naacl2022-ynidk,reddy:tacl2019coqa,choi:emnlp2018-quac,raina:acl2022-mcqnone}.
These are typically achieved by pairing one question with the context for another question. 
Introduction of unanswerability in the dataset in a controlled manner is significantly more challenging in KBQA, since the KB is the single shared context across questions and across train and test.



\section{Conclusions and Discussion}
\label{sec:conclusion}

We have introduced the task of detecting answerability when answering questions over a KB.
We have released GrailQAbility\footref{codefootnote} as the first benchmark dataset for KBQA with unanswerable questions, along with extensive experiments on three KBQA models. 
We find that no  model is able to replicate its answerable performance for the unanswerable setting even with appropriate retraining and thresholding, though both these methods of adaptation help in improving performance substantially.

Further, we find that there is a trade-off between robustness to missing schema elements and missing data elements.
The models find schema-level incompleteness easier to handle while data-level incompleteness substantially affects the models that enforce data-level constraints while decoding.
Another observation is that the models get quite confused for those unanswerable questions that contain a schema element seen in an answerable train question, along with a missing schema element that is not seen at training.
Finally, while TIARA turns out to be the best overall model, different models find different categories of unanswerability to be more challenging.
This suggests that new KBQA models will need to combine architectural aspects of different existing models to best handle unanswerability.
We believe that our dataset and observations will inspire research towards developing more robust and trustworthy KBQA models.

\section*{Acknowledgements}
Prayushi is supported by a grant from Reliance Foundation. Mausam is supported by grants by TCS, Verisk, and the Jai Gupta chair fellowship by IIT Delhi. He also acknowledges travel support from Google and Yardi School of AI travel grants. We thank the IIT Delhi HPC facility for its computational resources.

\section*{Limitations}\label{sec:limitations} 

Our dataset creation process - introducing unanswerability into a dataset of answerable KB questions by deleting KB elements - limits the nature of unanswerable questions.
All of these become answerable by completing the provided KB.
However, other kinds of unanswerability exists. 
Questions may involve false premise, for example, {\em C. Manning works at which European University?}, or may not even be relevant for the given KB.
We will explore these in future work.

Complete training and inference for each model with our dataset size takes 50-60 hours.
As a result, generating multiple results for the same models in the same setting was not possible and our results are based on single runs. 
However, using multiple runs with smaller dataset sizes we have seen that the variance is quite small.
Also, the dataset creation involves sampling KB elements for deletion and as such the generated dataset is one sample dataset with unanswerability.
This is unfortunately unavoidable when creating one benchmark dataset.

\section*{Risks}
Our work does not have any obvious risks.
In fact, addressing answerability reduces the risk of KBQA models confidently generating incorrect answers in spite of lack of knowledge.

\bibliography{anthology,custom}
\bibliographystyle{acl_natbib}

\appendix
\section{Appendix}

\begin{table*}[h]
        \begin{center}
            \small
            \begin{tabular}[b]{|c|ccc|ccc|cccc|cc|}

  \hline
 \multirow{3}{*}{Split} &  \multicolumn{3}{c}{Type Drop } &\multicolumn{3}{|c}{Relation Drop} & \multicolumn{4}{|c|}{Entity Drop}& \multicolumn{2}{|c|} {Fact Drop} \\ 
 & IID   &\multicolumn{2}{c|}{Z-Shot } & IID & \multicolumn{2}{c|}{Z-Shot} & \multicolumn{2}{|c}{IID} &\multicolumn{2}{c|}{Z-Shot } & IID & Z-Shot\\
 & NK   & P (NK)&F (NK) & NK & P (NK) & F (NK)& NA&NK & NA&NK & NA& NA  \\ \hline
Train &2667&0&0&2780&0&0&1288&1663&0&0&2952&0\\
Dev &211&154&47&211&146&55&91&89&87&151&176&241\\
Test &422&330&96&425&298&107&178&193&179&291&352&487\\
 \hline
            \end{tabular}

        \end{center}
        \caption{Statistics for unanswerable questions (P: Partial and F: Full) in GrailQAbility due to different types of KB incompleteness.}
        \label{tab:deatiled_dataset_description}
    \end{table*}

\begin{table}[h]
        \begin{center}
            \small
            \begin{tabular}[b]{|c|c|c|c|}

  \hline
 Split & IID   & Compositional & Z-Shot \\ \hline
Train & 23,933 & 0 & 0 \\
Dev & 1691& 488 & 1220\\
Test & 3386 & 981 & 2441\\
 \hline
            \end{tabular}

        \end{center}
        \caption{Statistics for answerable questions in GrailQAbility.}
        \label{tab:deatiled_dataset_description_answerable}
    \end{table}

\subsection{Details of Dataset Creation}
\label{subsec:details_of_datatset_creation}
In this section, we describe more details of the dataset creation process.



We assume the given KB to be the ideal KB $G^*$ and the given logical forms and answers to be the ideal answers $a^*$ and ideal logical forms $l^*$ for the questions $Q$.
We then create a KBQA dataset $Q_{au}$ with answerable and unanswerable questions with an `incomplete' KB $G_{au}$ by iteratively dropping KB elements from $G^*$.
Prior work on QA over incomplete KBs has explored algorithms for dropping facts from KBs~\cite{saxena:acl2020kgqaembed,thai:arxiv2022cbrikb}. 
We extend this for all categories of KB elements (type, relation, entity and fact) and explicitly track and control unanswerability.
At step $t$, we sample a KG element $g$ from the current KB $G_{au}^{t-1}$, identify all questions $q$ in $Q^{t-1}_{au}$ whose current logical form $l^{t-1}$ or path $p^{t-1}$ contains $g$, and remove $g$ from it.
Since $q$ may have multiple answer paths, this may only eliminate some answers from $a^{t-1}$ but not make it empty. If $g$ eliminate all answers from $a^{t-1}$, thereby making $q$ unanswerable.
If $q$ becomes unanswerable, we mark it appropriately (with $a^t=$\textsc{NA} or $l^t=$\textsc{NK}) and update $G_{au}^t=G_{au}^{t-1}\backslash\{g\}$.
This process is continued until $Q^t$ contains a desired percentage $p_u$ of unanswerable questions.

One of the important details is sampling KB element $g$ to drop.
In an iterative KB creation or population process, whether manual or automated, popular KB elements are less likely to be missing at any time.
Therefore we sample $g$ according to inverse popularity in $G^*$.
However, the naive sampling process is inefficient since it is likely to affect the same questions across iterations or not affect any question at all.
So, the sampling additionally considers the presence of $g$ for $Q_{au}^t$ --- the set of questions in $Q_{au}^t$ whose current logical form or answer paths contains $g$. Unlike schema elements, for selecting data elements to drop, we consider all data elements to be equally popular.

Next we describe how we drop all categories of KB elements in the same dataset.

\paragraph{Combining Drops:}
Our final objective is a dataset $Q_{au}$ that contains $p_u$ percentage of unanswerable questions with contributions $p_u^f$, $p_u^e$, $p_u^r$ and $p_u^t$ from the four categories of incompleteness.
Starting with the original questions $Q^*$ and KB $G^*$, we execute type drop, relation drop, entity drop and fact drop with the corresponding percentage in sequence, in each step operating on the updated dataset and KB.
For analysis, we label questions with the drop category that caused unanswerability.
Note that a question may be affected by multiple categories of drops at the same time.

GrailQA~\cite{gu:www2021grailqa} only contains the SPARQL queries for the questions (in English language) and the final answers, but not the answer paths.
To retrieve the answer paths, we modify the provided SPARQL queries to return the answer paths in addition to the final answer, and then execute these queries.
In Tab.\ref{tab:deatiled_dataset_description}, we include detailed statistics for unanswerable questions in GrailQAbility. We will release the GrailQAbility under the same license as GrailQA i.e.,  CC BY-SA 4.0. 


\begin{table*}[h]
\begin{center}
\small            
\begin{tabular}[b]{|c|l|ccc|ccc|ccc|}
\hline
  \multicolumn{1}{|c}{Train}&\multicolumn{1}{|c}{Model}&\multicolumn{3}{|c}{IID}
  
  & \multicolumn{3}{|c}{Compositional }  & \multicolumn{3}{|c|}{Zero-Shot}  \\

 & & F1(L) & F1(R) & EM & F1(L) & F1(R) & EM & F1(L) & F1(R) & EM \\ \hline
  \multirow{6}{*}{A}&RnG-KBQA &85.5 & 85.4  &83.2    &  65.9  &  65.9  &60.2& 72.7  & 72.7  &  67.3 \\ 
 &RnG-KBQA+T &79.0  &79.0  &77.3    &58.8  &58.8  &54.5&65.8  &65.8  &61.9  \\ 
  &ReTraCk &79.6  &79.5  &75.6    &63.1  &63.1  &55.4 &51.0  &51.0  &46.8  \\

   &ReTraCk+T &79.0  &78.9  &75.2   &61.6  &61.6  &53.9  &47.8  &47.8  &44.5 \\
    &TIARA & 88.9& 88.8 &86.8    &{\bf74.2}  &{\bf74.2}  &65.9 &{\bf78.1}  &{\bf78.1}  &{\bf73.9}  \\

   &TIARA+T & 84.1 & 84.0 &82.6   &67.7  &67.7  & 60.9 & 70.6 &70.6  &67.5 \\
  \hline 
  \multirow{6}{*}{A+U}&RnG-KBQA &  85.4  &85.3  & 83.3    &65.8  &65.8  &  60.8 &66.9  &66.9  &62.6 \\ 
 &RnG-KBQA+T &80.9  &80.9  &79.2   &60.5  &60.5  & 56.1&61.1  &61.1  &57.6   \\ 
  &ReTraCk &77.8  &77.6  &73.9    &60.6  &60.6  &53.5 &39.0  &39.0  &35.8  \\
  &ReTraCk+T &77.7  &77.5  &73.8 &59.9  &59.9  &52.8 &38.9  &38.9  &35.7 \\
  &TIARA & {\bf89.1}& {\bf89.0} &{\bf87.3}    &73.1  &73.1  &{\bf68.7} & 72.9 &72.9  &69.6  \\

   &TIARA+T & 85.5 &85.5  & 84.2  &66.3  &66.3  & 62.8 &66.8  &66.8  &64.2 \\
  \hline 
\end{tabular}
\end{center}
\caption{Performance of different models for answerable IID, compositional, and zero-shot test scenarios in GrailQAbility. Names have the same meanings as in Tab.~\ref{tab:mainresult}.}
\label{tab:test_scenarios_answerable}
\end{table*}

\subsection{Lenient Answer Evaluation} \label{sec:lenientevaluation}
Under lenient evaluation (L) for a given question, we calculate precision and recall w.r.t both gold answer in $Q_{au}$ and ideal answer in $Q$.
We consider maximum over $Q_{au}$ and $Q$ for precision and recall, and then calculate F1 as usual for calculating F1(L). 

\subsection{Model Adaptation and Training Details}
\label{sec:traindetails}

\paragraph{RnG-KBQA:} RnG-KBQA ~\cite{ye:acl2022rngkbqa}  consists of four modules: Entity Linker, Entity Disambiguation, Ranker and Generator. We use the same training objective and base models for re-training of these components on GrailQAbility. Similar to GrailQA ~\cite{gu:www2021grailqa}, for mention detection, we fine-tune a BERT-base-uncased model for 3 epochs with a learning rate of 5e-5 and a batch size of 32. For training the Entity Disambiguator, similar to RnG-KBQA, we fine-tuned a BERT-base-uncased ~\cite{Devlin2019BERTPO} model  for 3 epochs with a learning rate of 1e-3 and a batch size of 16. We use a non-bootstrapped strategy for sampling negative logical forms during the training of the ranker and fine-tune a BERT-based-uncased model  for 3 epochs with a learning rate of 1e-3 and a batch size of 2. As a generator, we fine-tune T5-base ~\cite{2020t5} for 10 epochs with a learning rate of 3e-5 and a batch size of 8. During inference with the generator, similar to RnG-KBQA, we use a beam size of 10 but due to the presence of NA questions in the test we do not perform execution augmented-inference. We compute the entity threshold $\tau_e$ and and logical form threshold $\tau_l$ based on disambiguation score and perplexity respectively by tuning on the validation set. During inference we use $\tau_e=-1.3890$ and $\tau_l=1.0030$ for RnG-KBQA A  and $\tau_e=-0.7682$  and $\tau_l=1.0230$ for RnG-KBQA A+U.

RnG-KBQA takes the (question, logical form) pair as input during training where the valid logical form also contains information about the mentioned entities in the question. We train two RnG-KBQA based KBQA models, one with answerable questions and the other with a combination of answerable and unanswerable questions. During training with A+U, we train mention detection and entity disambiguation model with questions having valid logic form i.e., $l=\mbox{NK}$, and perform entity linking for questions where $l\neq\mbox{NK}$. And Generator is trained to predict ``no logical form'' for unanswerable questions with $l=\mbox{NK}$ and valid logical form for remaining training questions. 


We use Hugging Face ~\cite{wolf-etal-2020-transformers}, PyTorch ~\cite{pytorch}  for our experiments and use the  Freebase setup specified on github \footnote{\url{https://github.com/dki-lab/Freebase-Setup}}. We use NVIDIA A100 GPU with 20 GB GPU memory and 60 GB RAM for training and inference  of RnG-KBQA on GrailQAbility which takes ~60 hours.

\paragraph{ReTraCk:} ReTraCk ~\cite{chen:acl2021retrack} includes three main components - retriever, transducer, and checker. Retriever consists of an entity linker that links entity mentions to corresponding entities in KB and a schema retriever that retrieves relevant schema items given a question. The entity linker has two stages - the first stage follows the entity linking pipeline described in ~\cite{gu:www2021grailqa} followed by a BOOTLEG ~\cite{orr2020bootleg} model used for entity disambiguation. We have used the pre-trained entity linker of ReTraCk.  We remove the dropped entities from the predictions of the entity linker.
The schema retriever leverages the dense retriever framework ~\cite{mazare2018training, humeau2019poly, wolf-etal-2020-transformers} for obtaining classes(types) and relations. Same as ReTraCk, we use pre-trained BERT-base-uncased model as a schema retriever and fine-tune it on GrailQAbility for 10 epochs with a learning rate of 1e-5. The best model is selected on basis of recall@top\_k where top\_k is 100 and 150 for types and relations respectively. We train two schema retriever models, one for A and one for A+U. For A, all answerable questions are used for training, while for A+U we use non-NK questions i.e. questions having only valid logical forms.

Transducer modules consist of a question encoder and a grammar-based decoder. ReTraCk uses a set of grammar rules for logical form. For NA training we have added a new grammar rule i.e. $\mbox{num} \rightarrow \mbox{NK}$ where NK is a terminal symbol representing No Knowledge. So for a question with no logical form, the sequence of grammar rules will be $\mbox{@start@}\rightarrow\mbox{num}$ and $\mbox{num}\rightarrow\mbox{NK}$. We have trained the transducer model with updated grammar rules for GrailQAbility. Training settings and hyperparameters are same as ReTraCk i.e. the BERT-base-uncased model with Adam optimizer and learning rate 1e-3, while learning rate for BERT is set to 2e-5. The best model is selected on basis of the average exact match calculated between predicted logical form and golden logical form.
Additionally, ReTraCk uses a Checker to improve the decoding process via incorporating semantics of KB. It consists of 4 types of checks i.e; Instance level, Ontology level, real and virtual execution. We have modified the stopping criteria for real execution. ReTraCk’s real execution terminates only when it finds a non-empty answer after query execution whereas we accept empty answers also after the execution of the query successfully (since unanswerable training involves empty answers). We compute the logical form threshold $\tau_l$ by tuning on the validation set. During inference we use $\tau_l=-6.5$ for ReTraCk A and $\tau_l=-7.5$ for ReTraCk A+U.
We use NVIDIA V100 GPU with 32 GB GPU memory and 60 GB RAM for the training of ReTraCk on GrailQAbility which takes ~50 hours. And we do inference on a CPU machine with 80GB RAM which takes ~3 hours.

\paragraph{TIARA:} TIARA ~\cite{shu-etal-2022-tiara} consist of four modules - Entity Retrieval, Schema Retriver, Exemplary Logical Form Retrieval and Generator. Entity Retrieval has three steps - mention detection, candidate generation, and entity
disambiguation. They have used there own mention detector called SpanMD. But since SpanMD is not open sourced so as suggested by authors we have used PURE mention detector which has similar performance to SpanMD. Candidates are generated using FACC1 ~\cite{facc1} and entity disambiguation pipeline is leveraged from ~\cite{ye:acl2022rngkbqa}. The logical form retrieval includes enumeration and ranking. It follows same methods as proposed in ~\cite{gu:www2021grailqa} and ~\cite{ye:acl2022rngkbqa}. So training process and hyper-parameters for this module is same as described in RnG-KBQA section above. Schema retrieval is implemented by a cross-encoder using pretrained BERT-base-uncased model. The model is trained for 10 epochs and best model is selected on the basis of recall@top\_k where k is 10 for both relations and classes. To train schema-retriever for A model we use all answerable questions while for A+U model we use questions with valid logical forms. Generator in TIARA takes following input - question, outputs of Entity Retrieval, Schema Retriver, Exemplary Logical Form Retrieval and outputs a logical form. Generation is performed by a transformer-based seq2seq model - T5(base) ~\cite{2020t5}. The Generator is fine-tuned for 10 epochs with learning rate 3e-5 and batch size of 8. We have trained two Generator models - A and A+U. For A model, all answerable questions are used for training, and for A+U model we use all answerable and unanswerable questions for training. For unanswerable questions the model is trained to generate output as "no logical form".
Similar to above models TIARA also performs beam search during inference with a beam size of 10. Additionally TIARA also performs constraint decoding to reduce generation errors on logical form operators and schema tokens. It uses a prefix trie to validate the sequence of tokens generated. After generation it is checked if the output is executable or not. Output is considered valid only if it executable (after constrained generation). Note: We consider executable queries with empty answers as valid query.

We use Hugging Face ~\cite{wolf-etal-2020-transformers}, PyTorch ~\cite{pytorch}  for our experiments and use the  Freebase setup specified on github \footnote{\url{https://github.com/dki-lab/Freebase-Setup}}.Training configurations for schema retriver are same as mentioned in ReTraCk and training configurations for Exemplary Logical Form Retrieval is same as mentioned in Rng-KBQA. We use NVIDIA A100 GPU with 40 GB GPU memory and 32 GB RAM for training TIARA Generator which takes around 8 hours for one model. Inference is performed parallely on 8 A100 GPUs with 40 GB GPU memory which takes around 1.5-2 hours.

\end{document}